\RequirePackage{fix-cm}
\documentclass[twocolumn]{svjour3}          
\smartqed  
\usepackage{graphicx}
\usepackage[numbers]{natbib}
\usepackage{wrapfig}

\usepackage{times} 
\usepackage{amsmath} 
\usepackage{amssymb}  
\usepackage{algorithm} 
\usepackage{algorithmic}
\usepackage{stmaryrd}

\usepackage{float}
\usepackage{subfig}
\usepackage{multicol}  
\usepackage{verbatim}
\usepackage{comment}

\usepackage{color}
\usepackage{url}

\newcommand{\edit}[1]{#1}
\newcommand{\editt}[1]{#1}

\usepackage{subfig}
\usepackage{multicol}
\usepackage{verbatim}

\usepackage{hyperref} 

\usepackage{fancyhdr} 
\usepackage{algorithmic}

\usepackage{algorithm} 

\usepackage[font=scriptsize]{caption}

\begin{document}

\title{
 Robots Learn Increasingly Complex Tasks with Intrinsic Motivation and Automatic Curriculum Learning
}
\subtitle{Domain Knowledge by  
 Emergence of Affordances, Hierarchical Reinforcement and Active Imitation Learning
}


\author{Sao Mai Nguyen \,$^{1,2,5}$        \and
        Nicolas Duminy\,$^{3,5}$ \and
        Alexandre Manoury\,$^{2,5}$ \and
        Dominique Duhaut\,$^{3,5}$ \and
        Cedric Buche\,$^{4,5}$ 
}


\institute{
$^{1}$ Flowers team, U2IS, ENSTA Paris, Institut Polytechnique de Paris \& Inria, France, 
$^{2}$ IMT Atlantique, Brest, France, 
$^{3}$ Universit\'e Bretagne Sud, Lorient, France, 
$^{4}$ ENIB, Brest, France, 
$^{5}$ Lab-STICC, UMR 6285, team RAMBO\\
\email{nguyensmai@gmail.com}           
}

\date{Received: date / Accepted: date}

\maketitle
\thispagestyle{fancy}
\lhead{}
\chead{
\texttt{\scriptsize{ Nguyen, S.M., Duminy, N., Manoury, A. et al. Robots Learn Increasingly Complex Tasks with Intrinsic Motivation and Automatic Curriculum Learning. Künstl Intell 35, 81–90 (2021). https://doi.org/10.1007/s13218-021-00708-8
 }}
\vspace{20pt}}
\rhead{}
\cfoot{}

\begin{abstract}

Multi-task learning by robots poses the challenge of the domain knowledge: complexity of tasks, complexity of the actions required, relationship between tasks for transfer learning. We demonstrate  that this domain knowledge can be learned to address the challenges  
in life-long learning. Specifically, the hierarchy between tasks of various complexities is key to infer a curriculum from simple to composite tasks.
 We propose a framework for robots to learn sequences of actions of unbounded complexity in order to achieve multiple control tasks of various complexity.  \editt{Our  hierarchical reinforcement learning framework, named SGIM-SAHT, offers a new direction of research, and tries to unify partial implementations on  robot arms and mobile robots.} We outline our contributions to enable robots to  map multiple control tasks to sequences of actions: representations of task dependencies, an intrinsically motivated exploration to learn task hierarchies, and active imitation learning. 
While learning the hierarchy of tasks, it infers its curriculum by deciding which tasks to explore first, how to transfer knowledge, and when, how and whom to imitate. 
\keywords{ Intrinsic motivation \and Continual learning \and Curriculum learning \and Transfer learning \and Multi-task learning \and Hierarchical reinforcement learning }

\end{abstract}

\section{Introduction}
In the mainstream approaches based on classical artificial intelligence and machine learning, robotic engineering approaches have made several valuable application-specific impacts. Yet, the achievements are often subject to restrictions that involve domain knowledge, a bounded and specific environment, or a limited set of tasks of the same complexity.

\begin{figure}
\centering

\minipage{.5\textwidth}
\minipage{0.49\textwidth}
  \includegraphics[width=.97\linewidth]{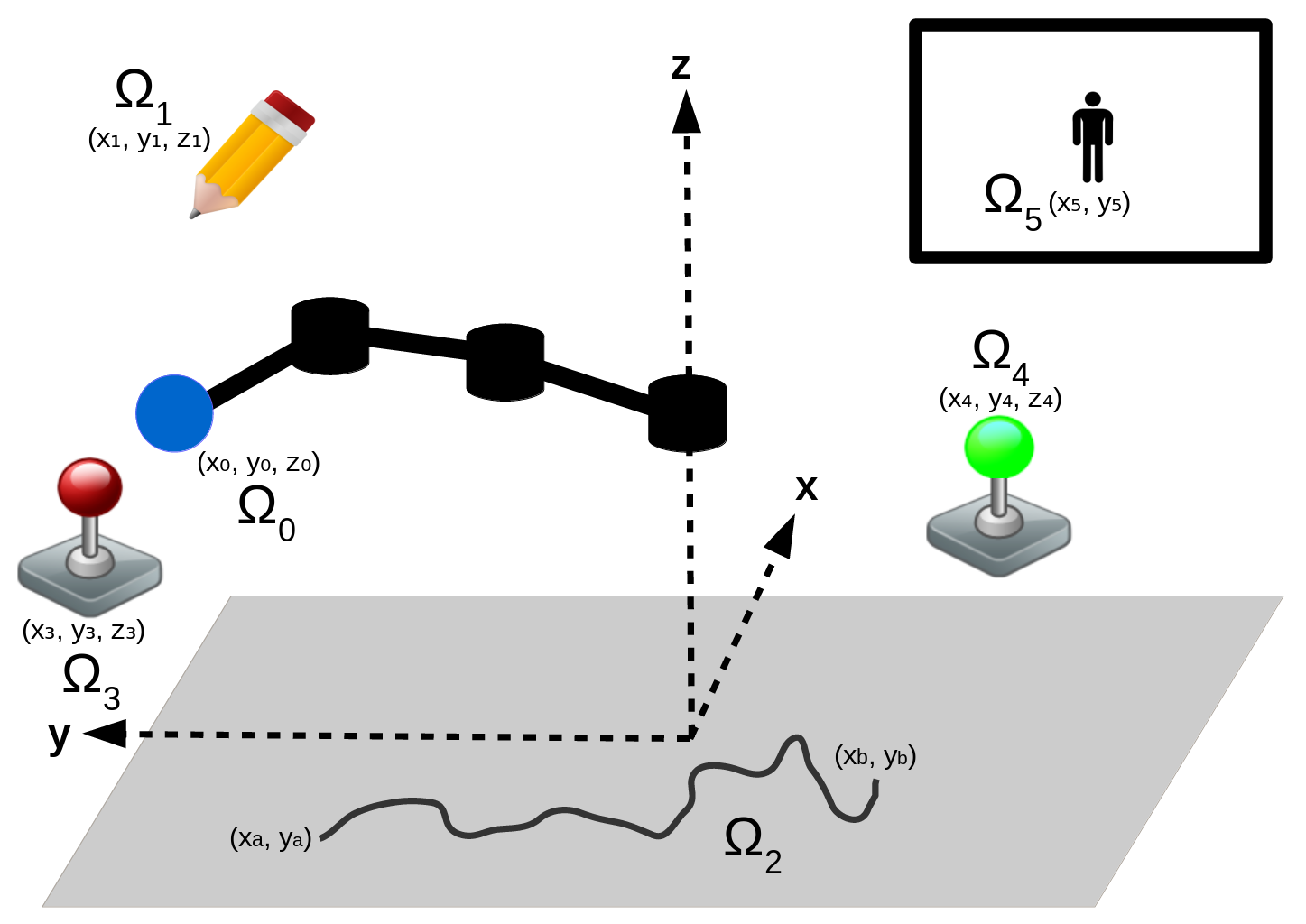}
\endminipage
\hfill
\minipage{0.49\textwidth}   
\includegraphics[width=.97\linewidth]{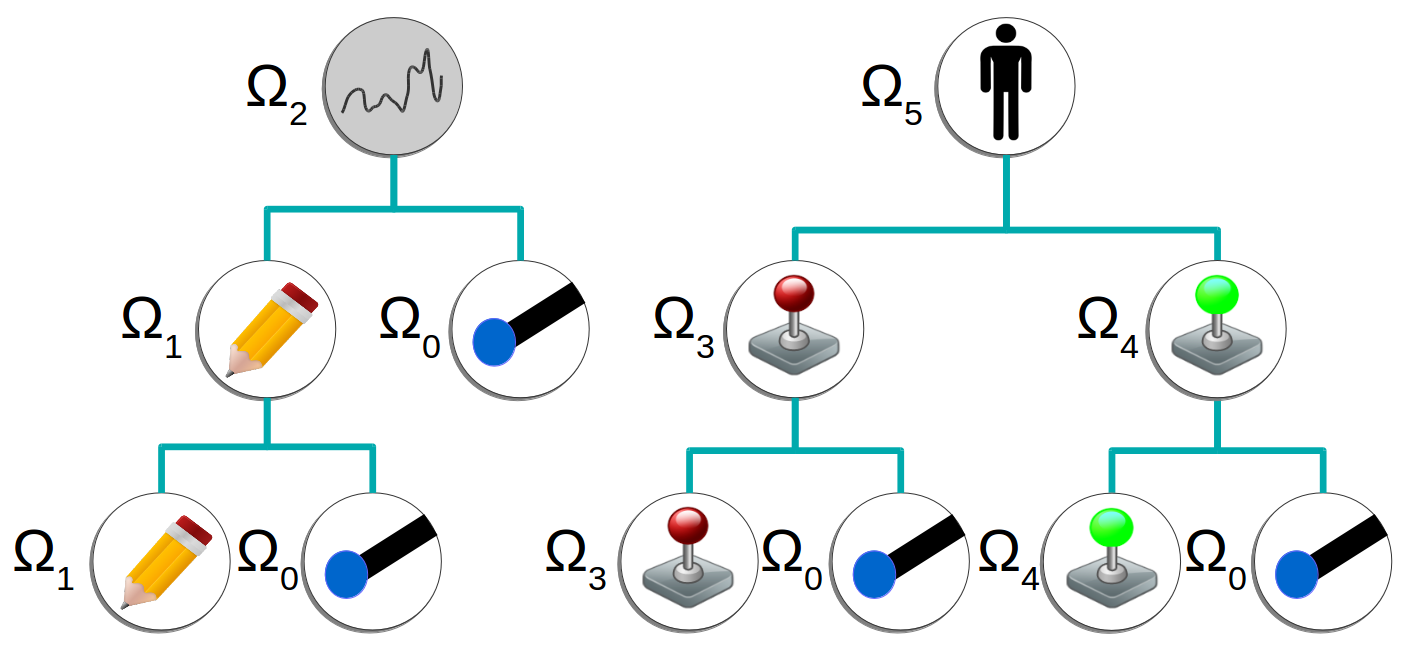}
 \endminipage  
\vspace{-0.2cm}
  \captionof{figure}{Setup1: a robotic arm, can interact with the different objects in its environment (a pen and two joysticks). Both joysticks enable to control a video-game character. The pen can be used to draw. Left: the simulation environment. Right: the corresponding hierarchy of models}
  \label{fig:setupSkaro}
\endminipage%
\\
\minipage{0.5\textwidth}
\minipage{0.49\textwidth}
  \includegraphics[width=0.97\linewidth]{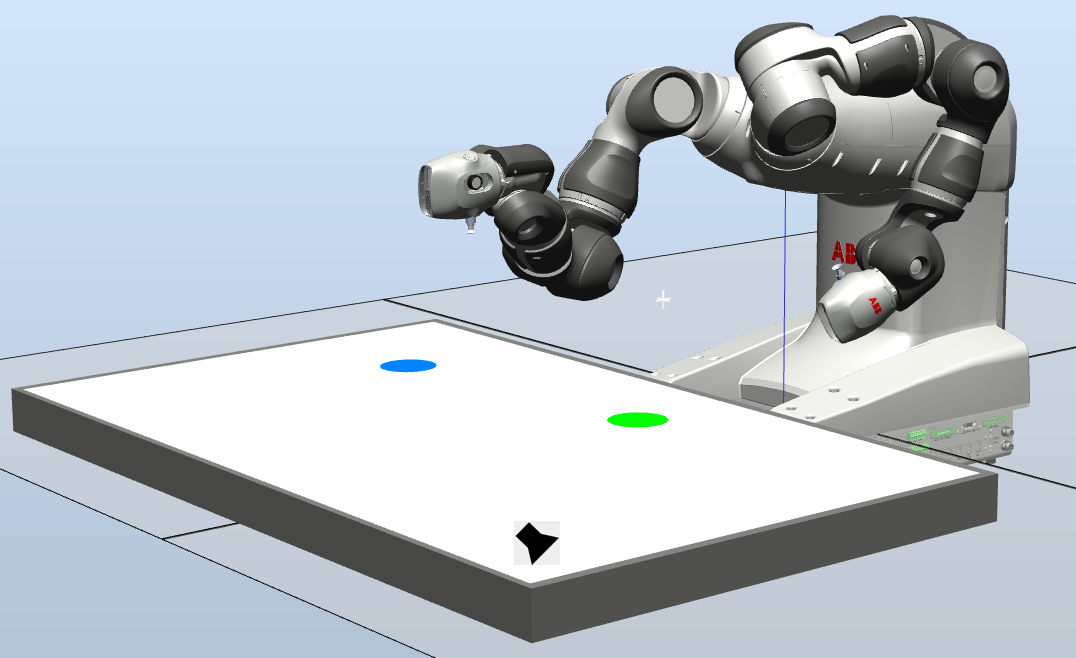}
\endminipage
\hfill
\minipage{0.49\textwidth}
\includegraphics[width=.97\linewidth]{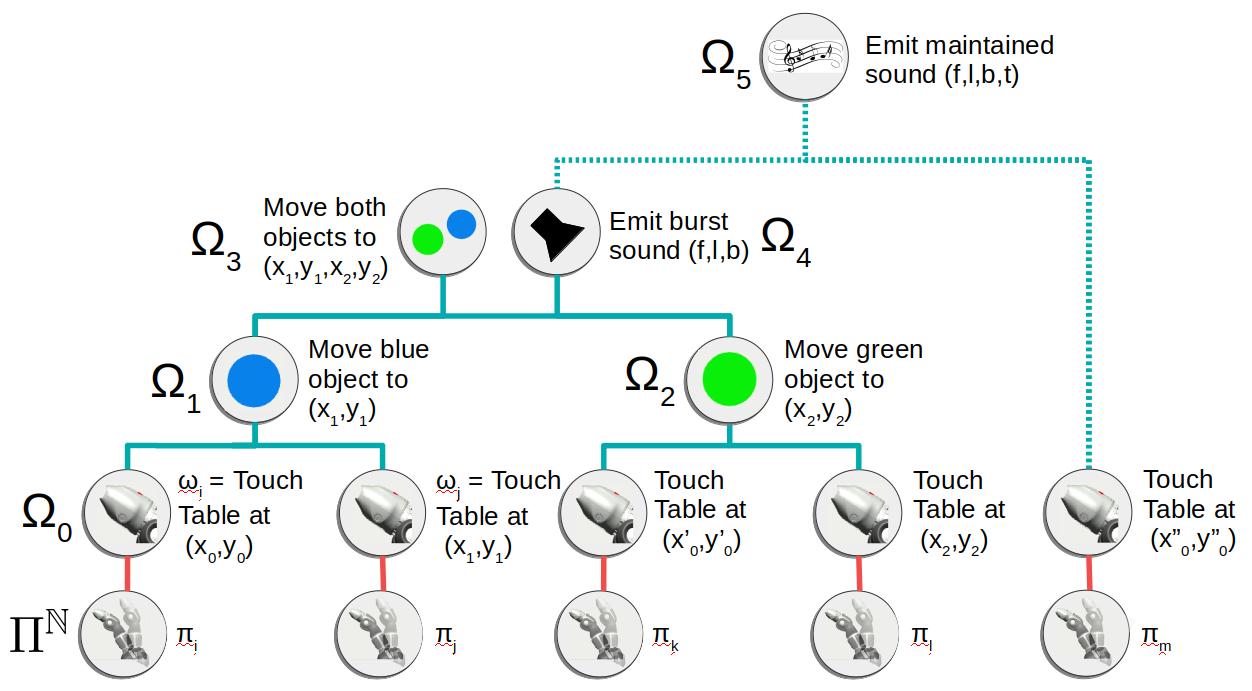}
 \endminipage  
\vspace{-0.2cm}
 \captionof{figure}{Setup2: the robot arm can produce sounds by moving the blue and green objects}
  \label{fig:setupYumi}
\endminipage
\\
\minipage{.5\textwidth}
\minipage{0.49\textwidth}
  \includegraphics[width=.97\linewidth]{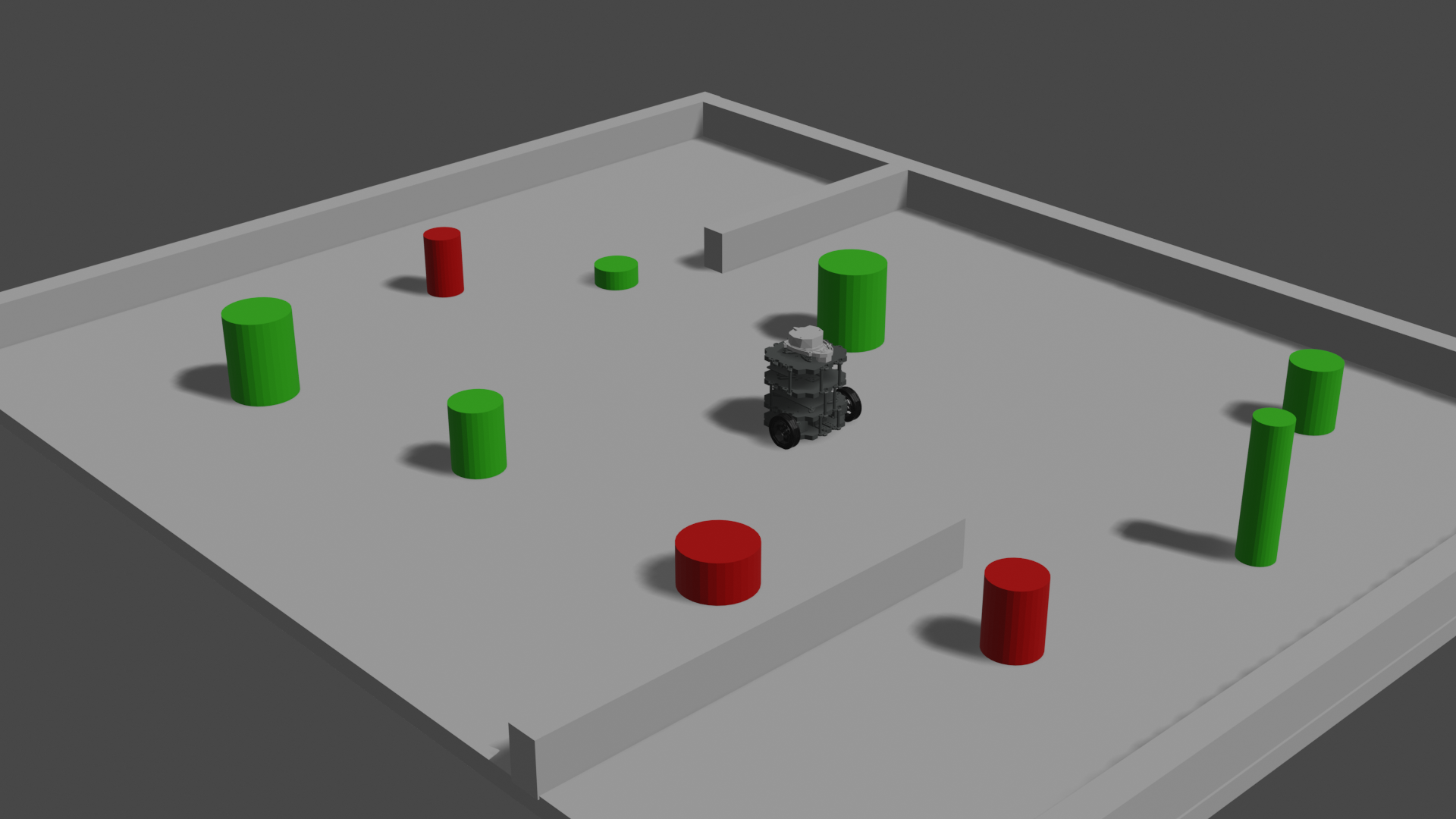}
\endminipage
\hfill
\minipage{0.49\textwidth}   
\includegraphics[width=.97\linewidth]{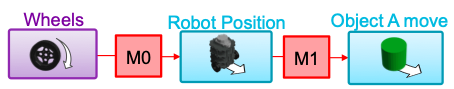}
\label{fig:nestedModels}
 \endminipage  
\vspace{-0.2cm}
   \captionof{figure}{Setup3: the mobile robot can avoid red obstacles, move green objects, push green objects with other objects. The objects are of random sizes.}
  \label{fig:setupChime}
\endminipage
\vspace{-0.7cm}
\end{figure}

To face the challenges of multi-task learning in a continual manner for embodied agents interacting with their stochastic environment and with humans, methods have taken inspiration in the living, and especially in how adults and infants learn as they develop, adapt and create new skills all along their lives.  
 These methods fall into the field named \textit{cognitive developmental robotics} \cite{Asada2001,Cangelosi2015}, \editt{within which we examine representations of actions and task relationships, their application to affordance learning and learning methods based on autonomous and socially guided exploration.}  

\subsection{Learning Sequences of Motor Policies}

In the case of multiple control tasks \edit{(ie. induced by actions)}  with various complexities and dimensionalities, the complexities of \edit{ the actions  required to complete them} should be unbounded, without a priori knowledge.  If we relate the complexity of actions to their dimensionality, actions of unbounded complexity should belong to spaces of unbounded dimensionality. For instance in setup Fig.\ref{fig:setupSkaro}, if an action of  dimensionality $n$ is sufficient to draw a letter, a sequence of 2 \edit{such} actions, i.e. an action of dimensionality $2n$ is sufficient to draw 2 letters. 
Generally, texts have variable lengths, thus there is no bound to the length of the sequence of actions. Likewise, an unbounded sequence of actions is needed to play tunes of any length in setup Fig.\ref{fig:setupYumi}; and to move all green objects side by side in setup Fig.\ref{fig:setupChime}.  
 Hence, here, we consider actions of unbounded complexity and suppose that they can be expressed as a sequence of action primitives. We consider \textit{action primitives} and \textit{sequences of actions}, also named in \cite{Zech2019IJRR} respectively \textit{micro actions} and \textit{compound actions}. The agent thus needs to estimate the complexity of the task and deploy actions of corresponding complexity.
 
 \editt{To tackle the curse of dimensionality which increases as sequences of actions grow longer}, methods such as \textit{DQN} \cite{mnih2015humanlevel}, using a deep neural network architecture have successfully handled high dimensional continuous outcome and context spaces \edit{but still with discrete action spaces}.  \citet{MnihBMGLHSK16} proposed an asynchronous  variant  of the \textit{Actor Critic} algorithm, relying both on deep neural networks and gradient policies. It successfully handles continuous action spaces, \edit{but still of predefined dimensionality}. 
The \textit{options} framework proposes a temporally abstract representation of actions \cite{Sutton1999AI},  \editt{ leading to sequences of actions for later reuse}. Approaches combining \textit{options}  with TD networks enable compositional prediction \cite{Sutton2006ANIPSN}. 
\editt{Learning simple skills and planning sequences of actions instead of learning a sequence directly has been proposed as \textit{Skill chaining}}  \cite{konidaris_barto2009_skill_chaining_learning}.  This forward chaining was successfully implemented to generate  multi-step plans for robot affordance learning \cite{ugur_piater2016_interdependant_affordance_learning_im}.

 Following the  ideas of a temporally abstract representation of actions and of multi-step planning, we propose a goal-directed representation of compound actions and a multi-step plan using inverse chaining of self-discovered subtasks.

\subsection{Task Hierarchy for Curriculum Learning} 
 \editt{This requirement of unbounded complexity of actions stems from the objective of learning unknown multiple tasks.} 
Multi-task learning in biological agents is progressive and continual. Humans and other species develop and create new skills all along their lives as they adapt to their environment and to their own needs. In particular, infants learn skills of increasing level of difficulty as they grow up: mastering simple skills first and then learning more complex skills based on the previous simple ones. Indeed, in multi-task learning problems, some tasks can be compositions of simpler tasks, which we call \textit{'complex tasks\'} or \textit{'composite tasks\'}. The learning agent should be \textit{starting small} before trying to learn more complex tasks, as phrased in \cite{Elman1993C}. Devising the order in which tasks should be learned has been coined \textit{'curriculum learning'} in \cite{Bengio2009P2AICML}: a learning agent needs to decide at each episode both which tasks it wants to learn to control (goal) and which actions to try (means). In our works, we thus take the hypothesis that tasks can be hierarchically related, some may be considered as subtasks of more complex tasks (\textit{'hierarchically organised tasks\'}). We conjecture that this hierarchy can help bootstrap the learning, by transfer of knowledge from simple to complex tasks.
 
Indeed, when given a task hierarchy, the robot can exploit this domain knowledge to reuse previously acquired skills to build more complex ones for tool use, as shown in \cite{Duminy2016I2JIICDLER}.
Approaches for hierarchical multi task learning with neural networks have also been proposed, such as \textit{Hierarchical DQN} \cite{Kulkarni2016ANIPS}, that uses intrinsic motivation to train a neural network in a fixed hierarchical manner.

To depend less on domain knowledge, the works presented in this article seek to learn the hierarchy between tasks, by exploring the different combinations between tasks. 
While learning the dependencies between tasks, we show in \cite{Duminy2019FN}, that reusing the knowledge of simple tasks as subgoals for more complex tasks indeed greatly reduced the exploration, and in \cite{Manoury2019HAI}, that planning can be used in combination of emerging hierarchical models of tasks.

\subsection{Emergence of Hierarchical Affordances}
An application case of motor learning and task hierarchy is affordances learning. The concept of \textit{affordance} has been first introduced by Gibson in \cite{Gibson1977PAK} to characterise physical states in an action-oriented fashion, in terms of the possible interactions an agent may have with objects. Even without knowing an object specifically, seeing visual cues of a handle may suggest possible embodied interactions.  

In affordances learning, many approaches have been developed \cite{Jamone2016ITCDS}:  for instance, the traversability affordance has been studied in different works \cite{Mitriakov2020WCCI}. Likewise, the grasp affordance is a recurrent topic and various approaches exist to learn it such as learning based on visual descriptors or raw image input \cite{5175529, Sergey2018}. However such methods focus on one, or a fixed number of specific affordances, with no mechanism adapting it to new or more complex affordances.

 We aim to continual learning of multiple affordances through the interaction with its environment. Thus, the robot builds sensory motor skills using a wide variety of actions. The robot can use actions of unbounded length and duration, in a continuous action space. 
 \citet{Ugur2016ITCDS} proposed an emergence of a hierarchical structure of affordances. However, affordances were defined as a list of discrete effects on objects, and the actions are manually coded. We would like to tackle continuous features of affordances and be able to learn compound actions in continuous action spaces. We extended their work with intrinsic motivation and planning. 
 
 \vspace{-0.6cm}
\subsection{Intrinsic Motivation as an Exploration Heuristic}
To allow multi-task learning, 
  developmental methods have transposed into algorithms the notion of intrinsic motivation, that has been outlined as a key mechanism for exploration \cite{Deci1985}.
These methods  use a reward function that is not shaped to fit a specific task but is general to all tasks the robot will face. Tending towards life-long learning, this approach, also called artificial curiosity, may be seen as a particular case of reinforcement learning using an intrinsic reward function.

Methods based on \textit{Q-Learning} and intrinsic motivation have been proposed in \cite{Vigorito2010} for discrete environments\editt{ where the reward depends on how much new information have been acquired.
Other methods used intrinsic motivation to explore the action space of the robot, based on empirical measures of prediction progress, such as algorithms \textit{IAC} \cite{oudeyer_kaplan2004_iac} and its active learning version \textit{RIAC} \cite{Baranes2009ITAMD}. Then \citet{Baranes2013RAS} added goal-babbling to explore the outcome space \edit{to address higher dimensional action spaces} with the \textit{SAGG-RIAC} algorithm. 
More recently, intrinsic motivation and goal babbling have combined deep neural networks and replay mechanisms \edit{for automatic curriculum learning of multiple tasks of different complexities}. IMGEP~\cite{Forestier2017C} -- a formalisation of unsupervised multi-goal RL -- GEP-GP~\cite{Colas2018ICMLI} -- mixing evolutionary methods and deep RL -- and CURIOUS~\cite{Colas2019P3ICML} -- mixing parametrised reward function and automated curriculum learning --  could select goals in a developmental manner from easy to difficult tasks. }

\subsection{Active Imitation Learning}
Methods taking advantage of human demonstrations have shown to tackle more varied and large goal spaces. They were combined with intrinsic motivation such as in  
  \cite{Nguyen2014AR}. 
The bootstrapping effect is all the more efficient when the learning robot uses active learning based on intrinsic motivation to choose what to learn, and who, when and how to imitate. This choice on the source of information, has been called a \textit{active imitation learning} \editt{and corresponds to the psychological description of infants' selectivity of social partners in~\cite{Begus2018ALFICSMCLMCuriousLearners:How}}. The SGIM-ACTS 
 algorithm proposed in~\cite{Nguyen2012PJBR} for multi-task learning has been applied to 3D object recognition 
   \cite{Nguyen2013IICDLE} or 
    mother tongue imitation by a vocal tract \cite{Moulin-Frier2014FP}.
 
\subsection{Summary and Position}
Grounding our studies in cognitive developmental robotics, we aim for a robot capable \textbf{of discovering and learning multiple tasks  as well as its curriculum by leveraging the relationships between tasks}. 
In this article, we show that domain knowledge about task relationship and complexity can be learned in order to adapt the complexity of the compound actions (sequences of actions)  required. \editt{The task relationships between known or emerging tasks can be discovered  with intrinsically motivated exploration or active imitation}.  We propose a common algorithmic architecture based on intrinsically motivated exploration to implement these mechanisms. 
 In the following sections, we present a formalisation of the problem of hierarchical learning, two representations of task hierarchy and a \editt{unifying algorithmic architecture with 3 partial implementations \cite{Duminy2018ICSC,Duminy2019FN,Manoury2019HAI}. We show how they} address the following questions:

\begin{itemize}
\vspace{-0.3cm}
\item how can knowledge from easier tasks be transferred to complex tasks while the robot learns tasks relationships?
\item how can a robot discover new learnable tasks from which to transfer knowledge to more complex tasks?
\item how can human demonstrations  be beneficial to an active learning robot tackling multiple control tasks?
\vspace{-0.3cm}
\end{itemize}
\editt{Compared to \cite{Duminy2018WCUSLI}, we have made the formalisation more coherent and have updated the implementations and experimental setups taken from \cite{Duminy2019FN,Manoury2019HAI}, and position these works better in comparison with existing state-of-the-art.
}

\section{Hierarchical Learning Framework}
 
In this section, we  propose a formalisation of the problem of learning compound actions to achieve hierarchically organised tasks, and propose an algorithmic architecture based on intrinsic motivation to learn the curriculum, \editt{common to the algorithms IM-PB, CHIME and SGIM-PB \cite{Duminy2018ICSC,Duminy2019FN,Manoury2019HAI}}.

\subsection{Formalisation}
\label{sec:formalisation}
Let us consider a robot interacting with a non-rewarding environment by performing sequences of motions of unbounded length in order to induce changes in its surroundings.

Each of these motions is named a \textit{primitive action}, described by a parametrised function with $p$ parameters: $a \in \mathcal{A} \subset \mathbb{R}^p$. We call $\mathcal{A}$ the \textit{primitive action space}. 
Our robot can perform sequences of primitive actions. Let a \textit{compound action} be a sequence of any length $n \in \mathbb{N}$ primitive actions, and be described by $n*p$ parameters : $a = [a_1, \dots, a_n] \in \mathcal{A}^n$. Thus the action space exploitable by the robot is a continuous space of infinite dimensionality $\mathcal{A}^{\mathbb{N}} \subset \mathbb{R}^{\mathbb{N}}$.

The actions performed by the robot have consequences on its environment, which we call outcomes $\omega \in \Omega$, where $\Omega$ is a subspace of the state space $S$ defining the control tasks to learn. Once the robot knows how to cause an outcome $\omega$, we say the outcome is then \textit{controllable}. The set of controllable outcomes is  $\Omega_{cont} \subset  \Omega$ and this set changes as the robot learns new tasks. For convenience, we define the \textit{controllable space}  $\mathcal{C} = \mathcal{A} \cup \Omega_{cont}$, regrouping both primitive actions $\mathcal{A}$ and observables that may be controlled, $\Omega_{cont}$.

The robot learns tasks/models $T$ each mapping controllable $c \in \mathcal{C}_T \subset \mathcal{C}$ and outcomes $\omega \in \Omega_T \subset \Omega$ within a given context $s \in S_T \subset S$. 
More formally, a task is a set of: a \textbf{forward model} $M_T:(S_T,\mathcal{C}_T) \to \Omega_T$ and an \textbf{inverse model} $L_T:(S_T,\Omega_T) \to \mathcal{C}_T $.  The forward model is used to predict the observable consequence ${\omega}$  from a given context $s$ of a controllable $c$, which is either a primitive action or a goal state that can be induced by a compound action. Conversely, the inverse model is used to estimate a controllable $\tilde{c}$ to be performed in a given context $s$ to induce a goal observable state $\tilde{\omega}$: if $\tilde{c}$ is a primitive action the robot executes the action, otherwise it sets $\tilde{c}$ as a goal state and infers the necessary actions using other inverse models. We note that this definition of task $T$ assumes that all $\omega \in \Omega_T$ can be realised independently from other observables, ie. the task dimensions of $\Omega_T$ do not interact with the rest of the state space in any way. In our works, we assume that the creation of the tasks (either predefined for IM-PB and SGIM-PB, or emerging for CHIME) ensures this coherence.

These models are trained on the data \editt{recorded by the robot along its exploration in its dataset $\mathcal{D}$. We define a strategy $\sigma$ any process for exploration. For instance, we consider autonomous exploration and imitation learning strategies. Formally, we define it as a data collection heuristics, based on the current data,} that outputs a set of triplets of initial state, action and outcome:\editt{ $\sigma : \mathcal{D} \mapsto\{ (s, a, \omega) \}$.}

\edit{Let us also note $\mathcal{H}$ the hierarchy of the models used by our robot. $\mathcal{H}$ is formally defined as a directed graph where each node is a task $T$ and its successors are the components of $ \mathcal{C}_T $.  As our robot learns this hierarchy, $\mathcal{H}$ varies along time}. Its representation is detailed in section \ref{sec:hierachyRepresentation}.

\subsection{Algorithmic Architecture}

\begin{figure}[t!]
\centering
\includegraphics[width=1\hsize]{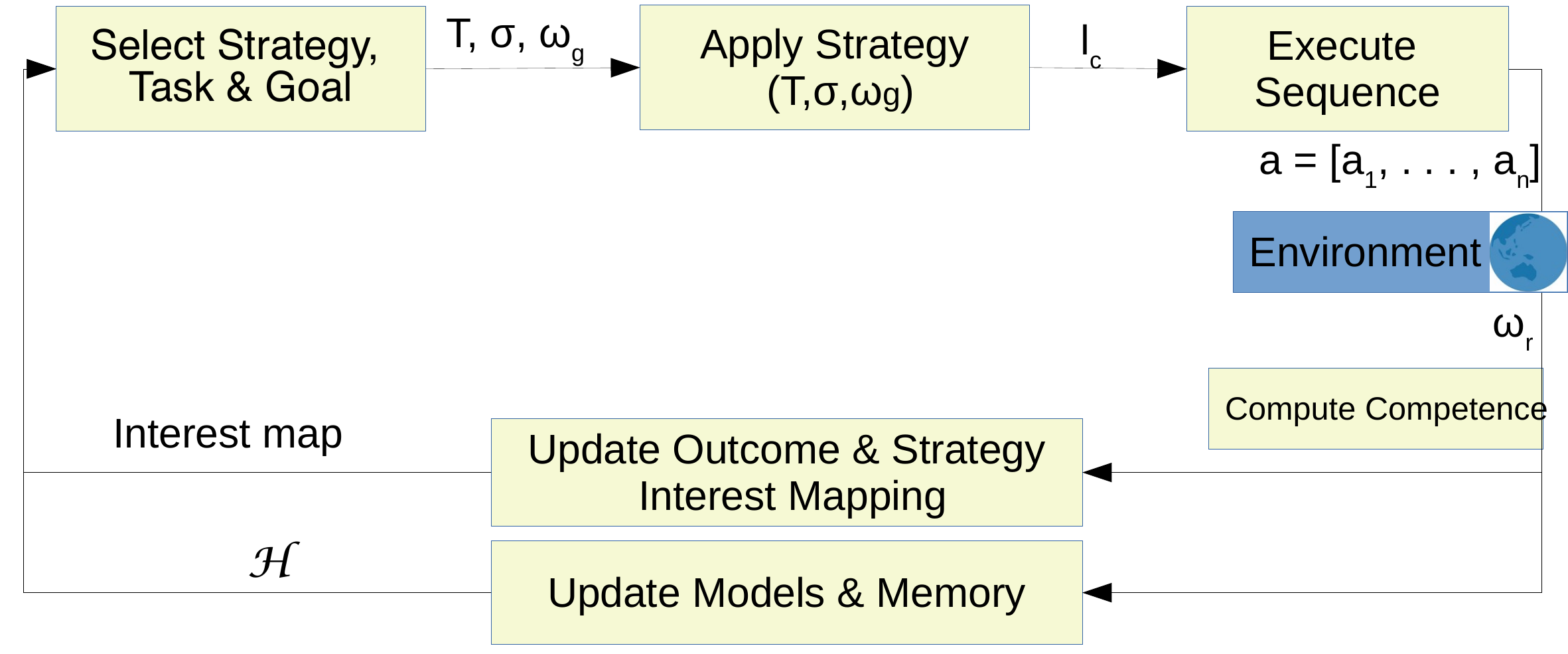}
\vspace{-0.7cm}
\caption{The SGIM-SAHT algorithmic architecture}
\vspace{-0.4cm}
\label{fig:algo}
\end{figure}

\editt{We describe} a generic algorithm Socially Guided Intrinsic Motivation for Sequence of Actions through Hierarchical Tasks (SGIM-SAHT) that learns and takes advantage of task hierarchy to solve increasingly complex tasks (Fig.\ref{fig:algo}, Alg.\ref{algorithm}).

\begin{algorithm}[tbh]
    \caption{SGIM-SAHT \label{algorithm}}
    \begin{algorithmic}[1]
        \REQUIRE the different strategies $\sigma_1,...,\sigma_n$
        \REQUIRE the initial model hierarchy $\mathcal{H}$
        \ENSURE partition of outcome spaces $R \gets \bigsqcup_i \lbrace\Omega_i\rbrace$
        \ENSURE episodic memory \textit{Memory} $\gets \emptyset$
        \LOOP
        \STATE $\sigma, T, \omega_g  \gets$ Select Strategy, Task \& Goal Outcome($R, \mathcal{H}$) \label{algo:strategy}
        \STATE $l_c \gets$ Apply Strategy($\sigma, \omega_g$) \label{algo:sequence}
        \STATE $\mathcal{D} \gets  (\omega_r, a, l_c)  \gets$ Execute Sequence($l_c$) \label{algo:memory}
        \STATE  $(comp(\omega_g), comp(\omega_r)) \gets $ Compute Competence$(\omega_g,\omega_r))$\label{algo:competence}
        \STATE Update $M_T, L_T, \mathcal{H}$ with $(\mathcal{D}, comp(\omega_g), comp(\omega_r))$ \label{algo:M}
        \STATE $R_i \gets$ Update Outcome and Strategy Interest Map($R,\mathcal{D},\omega_g$)
        \ENDLOOP
    \end{algorithmic}
\end{algorithm}

SGIM-SAHT learns by episodes in which a task $T$ to work on, a goal outcome $\omega_g \in \Omega_T$ and a strategy $\sigma$ have been selected to optimize progress and according to an interest map (see below).
The selected strategy $\sigma$  applied to the chosen goal outcome $\omega_g$ chooses a sequence of controllables $l_c=  [c_1, \dots, c_m] $ \edit{as a candidate to} reach the  $\omega_g$ (Alg.\ref{algorithm}, l.\ref{algo:sequence}). 

\editt{SGIM-SAHT chooses an adequate task $T_i$ , i.e. when the input space of $L_{Ti}$ includes $ (s,\omega_g)$. Then it} applies $L_{Ti}$ to find the action $a^i=L_{Ti}(c_i)$. This inference process may be recursive until the output is an action, using the hierarchy between tasks.  SGIM-SAHT thus infers from $l_c $  a compound action $a = [a_1, \dots, a_n] \in \mathcal{A}^{\mathbb{N}}$, to be executed by the robot. 
The trajectory of the episode with the primitive actions and controllables sequence and \editt{goal and reached outcomes $\omega_g,\omega_r$ are recorded in the memory}
 (Alg.\ref{algorithm}, l.\ref{algo:memory}).

 Then, it computes the learner's competence on the goal outcome. In the RL framework, this competence can be seen as the reward for the goal outcome. In our multi-task learning setting, we use \editt{as competence a  reward function common to all goals} based on the Euclidean distance between the goal outcome $\omega_g$ and the reached outcome $\omega_r$ (Alg.\ref{algorithm}, l.\ref{algo:competence}). \edit{Variations of this metric have been implemented in} \cite{Duminy2019FN,Manoury2019HAI}. 
 
The memory and competence are used to update the models $M_T$ and $L_T$, \edit{the set of tasks,} and the hierarchy of models $\mathcal{H}$ (Alg.\ref{algorithm}, l.\ref{algo:M}). \edit{These mechanisms differ in our 3 algorithms.} 
\editt{Besides, the competence is used to obtain an interest map that associates to each strategy and region of outcome space partition an interest measure to guide the exploration. The interest measure is computed as the progress or derivative of the competence of the enclosed outcomes (details in \cite{Nguyen2012PJBR}).

When the number of outcomes added to a region  $R_i$ exceeds a fixed limit, the region is split into two regions with a clustering boundary that separates outcomes with low from those with high interest (details in \cite{Nguyen2012PJBR}).}

\begin{table}[tb]
\begin{tabular}{|p{0.8cm} |p{2.8cm} |p{2cm} |p{1.4cm} |}
\hline
Algo. & Strategies $\sigma$ & Tasks set & Comp. Act. \\
\hline
IM-PB & Outcomes, Procedures explo. & Static set & Procedures \\
CHIME & Action space, outcome space explo. & Dynamic set (emerging tasks) & Planning \\
SGIM-PB &  Outcomes, Procedures explo.; active imitation & Static set & Procedures\\
\hline
\end{tabular}
\vspace{-0.2cm}
\caption{Differences between the 3 implementations of SGIM-SAHT}
\label{tab:DiffAlgo}
\vspace{-0.5cm}
\end{table}

\editt{
SGIM-SAHT has three implementations described in the sections \ref{sec:taskRelationship} and \ref{sec:interactive}}. Their differences are described in table \ref{tab:DiffAlgo}.

\section{Task Hierarchy Representation}

 \label{sec:hierachyRepresentation}

In this section, we describe two representations of task hierarchy,  called Procedure and CHIME, to be used to learn the domain knowledge about the relationship between tasks.

\subsection{Procedure Hierarchy Representation}

The first is a goal-directed representation of action sequences in the form of sequences of subgoals \edit{or task decomposition, that enable transfer of knowledge between inter-related tasks. It is a temporally abstract representation of a succession of actions.  
 More formally, \textbf{a procedure is defined as a  \edit{succession of outcomes}} $(\omega_1, \omega_2,...,\omega_n)$
. The succession is unbounded. The procedure space is $\Omega^{\mathbb{N}}$.  A task decomposition into a procedure is an association of a goal outcome to a procedure. 
Thus, an outcome represents a task node in $\mathcal{H}$, while the task decomposition represents the directed edges and the procedure is the list of its successors. $\mathcal{H}$ is initialised as a densely connected graph, and the exploration prunes the connexions by testing which procedures or task decompositions respect the ground truth.}

\begin{figure}[h!]
\vspace{-0.6cm}
\centering
\includegraphics[width=0.45\linewidth]{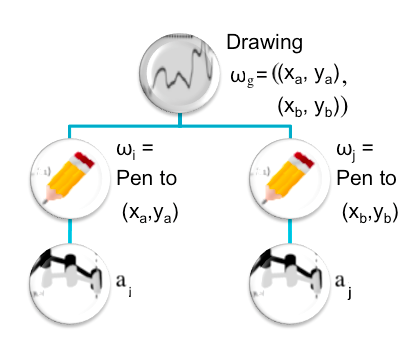}
\vspace{-0.3cm}
\caption{Illustration of a procedure for setup fig.\ref{fig:setupSkaro}. To make a drawing $\omega_g$ between points $(x_a, y_a)$ and $(x_b, y_b)$, a robot can recruit subtasks consisting in ($\omega_i$) moving the pen to $(x_a, y_a)$, then ($\omega_j$) moving the pen to $(x_b, y_b)$. These subtasks will be completed respectively with  actions $a_i$  and $a_j$. \edit{To complete this drawing, the learning agent can use the sequence of actions $(a_i,a_j)$} 
}
\label{fig:hierarchy}
\vspace{-0.6cm}
\end{figure}

Executing a procedure $(\omega_1, \omega_1, ..., \omega_n)$ means building the action sequence $a$ corresponding to the succession of actions $a_i, i \in \llbracket 1, n \rrbracket$ (potentially action sequences as well) and execute it (where the $a_i$ reach best the $\omega_i$ $\forall i \in \llbracket 1, n \rrbracket$ respectively). Fig.\ref{fig:hierarchy} illustrates this idea of task hierarchy. 

The procedures representation is based on a static set of \edit{(ie. predefined) }controllable and outcome spaces.

\subsection{CHIME Hierarchy Representation}
In comparison, the CHIME hierarchical representation is based on \textbf{a dynamic set of controllable and outcome spaces and the emergence of nested models}. It is built using simple models  \editt{$M_T:(S_T,\mathcal{C}_i) \to \Omega_j$} which can rely on others: lower models map outcomes to actions while higher models map them to other outcomes that should be reached. For instance in the setup3 (Fig.\ref{fig:setupChime}), the robot can move itself with the model ($ M_0:wheelCommand  \in \mathcal{A} \to (x_0,y_0) \in \Omega_0$) and it can learn that the position $(x_1,y_1)$ of the object pushed depends on its own position ($M_1:(x_0,y_0) \in \Omega_0 \to (x_1,y_1) \in \Omega_1$). The combination of the nested models 
 enable the robot to control $(x_1,y_1)$ with its wheel commands (Fig.\ref{fig:setupChime} Right
). In the hierarchy of models $\mathcal{H}$
, the models $M_T$ represent the directed graph from $\Omega_j$ to $\mathcal{C}_i$. 
At the beginning $H = \emptyset$, no model is present, and the robot chooses itself what model to create or modify: if $\mathcal{C}_i$ seems to be highly correlated to $\Omega_j$ it may create the model $M:\mathcal{C}_i \to \Omega_j$, which becomes the first nodes and edge of $\mathcal{H}$.

 \edit{\section{Learn Task Relationships by Autonomous Exploration}}
 \label{sec:taskRelationship}
The idea of SGIM-SAHT is to use a hierarchical representation of the outcome and controllable spaces. \edit{This representation outlines the dependencies between tasks in order to reuse previous knowledge and use actions of adapted complexity. In this section, we examine how task hierarchy can be learned in the cases of a static and a dynamic set of tasks.}

 \subsection{Learn Task Hierarchy from a Static Set of Tasks}
\edit{Let us consider that the set of tasks is given. The robot needs to choose which are easier to learn first, which are more difficult, and which tasks can be reused as sub-goals for more difficult models.} To learn the hierarchy $\mathcal{H}$ between tasks, we proposed in  \citep{Duminy2018IIRC} an implementation, IM-PB (Intrinsically Motivated Procedure Babbling) based on procedures and  the identification among all possible dependencies, of those that are valid.

\editt{IM-PB has 2 strategies : autonomous exploration of the outcome or the procedure spaces. Intrinsic motivation guides the exploration of the the task space and the procedure space (Alg.\ref{algorithm}, l.\ref{algo:strategy})} to find a curriculum from simple to complex tasks.  
IM-PB takes advantage of the dependencies between tasks : when executing a sequence \editt{(Alg.\ref{algorithm}, l.\ref{algo:sequence}), carrying out a procedure $(\omega_1,... \omega_n)$ means carrying out the action primitive sequence $\pi$ by executing sequentially each component action $a_i$, where $a_i$ is an action that reaches $\omega_i$ $\forall i \in \llbracket 1, n  \rrbracket$}. 

\editt{The experiments on setups of Fig.\ref{fig:setupSkaro},\ref{fig:setupYumi} in \cite{Duminy2018ICSC,Duminy2019FN} show that an intrinsically motivated learner is capable of learning sequences of motor actions. }During the learning phase, results show that the robot explores mainly the task space for simple tasks, and it explores considerably more the procedure space for complex hierarchical tasks. Thus it implicitly understands that simple tasks do not need to be decomposed into subtasks and can be reached directly by action primitives. On the contrary for complex tasks, it is more advantageous to seek which subtasks to reuse.
\editt{During the test phase, the robot uses the correct task decomposition}. Furthermore, It can also adapt the length of its action sequence to the task to achieve: the results show that the length of the sequence of actions increases as the complexity of the task increases in terms of its hierarchy. 
\textbf{Combining these procedures with the learning of simple actions to complete simple tasks, it can build sequences of actions to achieve complex tasks}.
We showed in \cite{Duminy2019FN} that the robot can take advantage of the procedures representation to improve its performance, especially on high-level tasks. It also adapts the complexity of its action sequence to the complexity of the task at hand.

Nevertheless, this adaptation is limited to the first two levels of task hierarchy, and the learner can not well adapt this complexity to a deeper hierarchy of tasks. To help the robot improve its understanding of task dependencies, we present in section \ref{sec:interactive} the benefits of active imitation learning.

\subsection{Learn Task Hierarchy from a Dynamic Set of Tasks}

In the previous section, the set of possible tasks (association of inputs and outputs) are given, and the robot needs to learn the relationship between them. In other terms, in $\mathcal{H}$, the nodes are pre-defined and the robot discovers the connections in the graph. 
 In this section, we consider the case where the robot needs to learn both the connections and the nodes, ie. task hierarchy at the same time as task emergence. 
  
 \subsubsection{Experimental Setup}
 It was applied to a wheeled robot with obstacles and movable objects of random size (Fig.\ref{fig:setupChime}). It proposed an emergence of affordances : the algorithm is able to discover learnable models \edit{(eg. move a green object)}, and once the model is learned, its output can be used as input features of more complex models \edit{(eg. push an object with a green object)}, leading to hierarchical learning. 
 In this paper, we describe an affordance as a task $T$.

\subsubsection{CHIME Implementation}

Let us consider that we do not have a set of sub-goals given, but have only given a high-dimensional set of inputs and observable outputs. \edit{ At initialisation, the robot only has access to the control of its wheels, the positions and physical properties of objects. The set of tasks and controllable outcomes is empty. By exploration, it discovers changes in its environments (eg. green object moved) and how it can control them (eg. its own position), but also how these changes can induce more complex outcomes (eg. push an object with another).}

The algorithm CHIME \citep{Manoury2019HAI}  implements SGIM-SAHT for a dynamic set of tasks and discovers new tasks by enactive exploration. At each episode,  the physical properties such as colour, height, diameter of objects are generated randomly. 
The main characteristics of CHIME, detailed in  \cite{Manoury2019HAI}, lie in its execution of sequence and its update of models and memory (Alg.\ref{algorithm}, l.\ref{algo:memory} and \ref{algo:M}).

To execute a sequence of controllables $l_c$ \editt{(Alg.\ref{algorithm}, l.\ref{algo:memory})}, for each element $c_i$ :
\begin{itemize}
    \item if $c_i$ is a primitive action, it is directly executed 
    \item if $c_i$ is not a primitive action, $c_i \in \Omega_{cont}$. An affordance $T(S_T,\mathcal{C}_T, \Omega_T)$ is then selected (with $c_i \in \Omega_{T}$) and its inverse model is applied to obtain the controllable $b_i = L_T(c_i)$. If $c_i$ is out of reach within a timestep, a planning phase is used to build a sequence of element of $\mathcal{C}_A$ in order to reach $c_i$. The same mechanism is applied recursively on it until having only primitive actions.
\end{itemize}

To update its models \editt{(Alg.\ref{algorithm}, l.\ref{algo:M})}, 
at the end of each episode, subspaces of $\Omega$ for which outcomes $\omega$ has been observed are listed. Then \editt{the robot verifies if  $\Omega$ matches a known affordance. To save computing time, we only verify for randomly selected subspaces of $\Omega$. 
If it does not match}, it  
 creates a new affordance, ie.  a forward and an inverse model. If it matches, but the new data contradicts with previous data (the competence for this task is reduced), it tries to update the model by adding context spaces.

\subsubsection{Developmental Emergence of Affordances}

The results on setup3 (Fig.\ref{fig:setupChime}) show  in \cite{Manoury2019HAI} that CHIME  discovers new affordances, and uses unbounded sequences of learned actions to complete all tasks. Even without predefinition of possible tasks
, the agent is able to discover the inputs and outputs of models of learnable tasks, and how they are related to each other.  We show in \cite{Manoury2019HAI} that these tasks emerge and are learned in a developmental order from the lowest to the highest level of hierarchy.  Planning based on these emergent tasks enables the robot to infer a sequence of actions to complete complex tasks. 
 
 The learning is based on active learning to collect data through new interactions with the environment, guided by the heuristics of intrinsic motivation. Once learned, these affordance control models are used to plan complex tasks with known or unknown objects, by using their physical properties to decide whether a learned affordance may be applied.
 
 \edit{
\subsubsection{Automatic Curriculum Learning}
Both IM-PB and CHIME rely on a temporally abstract representation of task relationships, whether all tasks have been predefined in advance or the robot has to discover new tasks. Discovering the task relationship enabled the robot to infer domain knowledge about the complexity of tasks and of the actions needed, but also to build its curriculum, learning simple, then increasingly complex tasks.
 }
 

\section{Who, What, How to Imitate to Learn Task Relationships}
\label{sec:interactive}
Beyond mere autonomous intrinsically motivated exploration, we show that \edit{domain knowledge can also be learned through social guidance}. In high dimensional and unbounded task spaces, the performance is improved even more when the robot can imitate actions and procedures from teacher demonstrations, when it actively chooses between self-supervised intrinsic motivation and imitation learning strategies.

For hierarchically organised tasks,  
 we proposed in \cite{Duminy2019FN,Duminy2018ICSC} the implementation SGIM-PB (Socially Guided Intrinsic Motivation by Procedure Babbling) that merged IM-PB with SGIM-ACTS. \editt{In addition to the strategies $\sigma$ of IM-PB, SGIM-PB has two strategies per teacher it can interact with : request a demonstration of actions or procedures. 
With the strategy \textit{mimicry of an action}, SGIM-PB requests a demonstration of an action to reach  $\omega_g$. To apply the strategy (Alg.\ref{algorithm}, l.\ref{algo:sequence}), SGIM-PB adds noise to the demonstrated action parameters to  explore locally the action parameters space, before executing the action (Alg.\ref{algorithm}, l.\ref{algo:memory}). 

With the strategy \textit{mimicry of a procedure}, the learner requests a procedure for  $\omega_g$. SGIM-PB executes the demonstrated procedure $(\omega_{di}, \omega_{dj})$ by adding noise to the parameters, thus exploring locally the procedure space.
 
 Using the competence measures and the interest map, SGIM-PB chooses when imitation learning is} more beneficial than autonomous exploration, who among the different teachers are most expert in the field of knowledge it needs at the moment, and what kind of demonstrations is most beneficial. In terms of imitation learning, SGIM-PB self-determines who, what and when to imitate. 
Through setups Fig.\ref{fig:setupSkaro},\ref{fig:setupYumi}, we show in  \cite{Duminy2019FN,Duminy2018ICSC} that demonstrations of procedures bootstrap the learning for tasks of the highest level of hierarchy. Moreover, demonstrations seem the most beneficial when they are demonstrations of policies for simple tasks, and when they are indications of procedures (i.e. subtasks) for complex tasks.


\section{Conclusion}

\begin{table}[]
{\small
\begin{tabular}{|l|l|l|l|l|l|l|}
\hline
Algorithm                                                                                                                                       
& \rotatebox[origin=c]{-70}{Reward}                 & \rotatebox[origin=c]{-80}{\begin{tabular}[c]{@{}l@{}}  Cont. goal\end{tabular}} & \rotatebox[origin=c]{-80}{Multi-task} & \rotatebox[origin=c]{-80}{\begin{tabular}[c]{@{}l@{}}Hierarchy\end{tabular}} & \rotatebox[origin=c]{-75}{Actions} & \rotatebox[origin=c]{-80}{Imitation}                                                \\ \hline
\begin{tabular}[c]{@{}l@{}} Qlearning  \cite{Sutton1998}      \end{tabular}                                                            & Ext.                                                             &                                                                  &           &                                                                 & Discrete                                                               &                                                           \\ \hline
\begin{tabular}[c]{@{}l@{}}Qlearning \& \\curiosity  \cite{Vigorito2010}\end{tabular}                                         & Int.        &                                                                  &           &                                                                 &                                      Primitive                          &                                                           \\ \hline
\begin{tabular}[c]{@{}l@{}}\editt{Option TDNet\cite{Sutton2006ANIPSN}}\end{tabular}                                            & Ext.                                                             &                                                                  &           &                                                                 & Option                                                            &                                                           \\ \hline

\begin{tabular}[c]{@{}l@{}}Skill- \\ chaining  \cite{konidaris_barto2009_skill_chaining_learning}\end{tabular} & Ext.                                                             &                                                                  &           & Yes                                                             & Option                                                            &                                                           \\ \hline
DQN \cite{mnih2015humanlevel}                                                                                                  & Ext.                                                             &                                                                  &           &                                                                 &       Discrete                                                         &                                                           \\ \hline

\begin{tabular}[c]{@{}l@{}} h-DQN \cite{Kulkarni2016ANIPS}     \end{tabular}                                                                                      & Int.        &                                                                  & Yes       & Yes                                                             &      Primitive                                                          &                                                           \\ \hline
\begin{tabular}[c]{@{}l@{}} Asynch. \\Actor Critic \cite{MnihBMGLHSK16}  \end{tabular}                                                                                            & \begin{tabular}[c]{@{}l@{}}Ext. \\ self-eval\end{tabular} &                                                                  &           &                                                                 &                Primitive                                                &                                                           \\ \hline
GEP-PG  \cite{Colas2018ICMLI}                                                                                                     & Int.        & Yes                                                              & Yes       &                                                                 &                                                                &                                                           \\ \hline
\begin{tabular}[c]{@{}l@{}} CURIOUS  \cite{Colas2019P3ICML}  \end{tabular}                                                                                                     & Int.        & Yes                                                              & Yes       &                                                                 &     Primitive                                                           &                                                           \\ \hline
\begin{tabular}[c]{@{}l@{}} IAC\cite{oudeyer_kaplan2004_iac}, RIAC\cite{Baranes2009ITAMD}  \end{tabular}                                                        & Int.        & Yes                                                              & Yes       &                                                                 &        Primitive                                                        &                                                           \\ \hline
\begin{tabular}[c]{@{}l@{}} SAGG-RIAC \cite{Baranes2013RAS} \end{tabular}
                                                                            & Int.        & Yes                                                              & Yes       &                                                                 &        Primitive                                                        &                                                           \\ \hline
\begin{tabular}[c]{@{}l@{}}IMGEP \cite{Forestier2017C} \end{tabular}                                          & Int.        & Yes                                                              & Yes       &                                                              &      Primitive                                                          &                                                           \\ \hline
\begin{tabular}[c]{@{}l@{}} SGIM-ACTS \cite{Nguyen2012PJBR}    \end{tabular}                                                                                             & Int.        & Yes                                                              & Yes       &                                                                 &     Primitive                                                           & Yes \\ \hline
\begin{tabular}[c]{@{}l@{}} IM-PB  \cite{Duminy2018IIRC,Duminy2019FN}    \end{tabular}                                                                                                & Int.        & Yes                                                              & Yes       & Yes                                                             & Proced.                                                    & \begin{tabular}[c]{@{}l@{}} \end{tabular} \\ \hline
\begin{tabular}[c]{@{}l@{}} SGIM-PB \cite{Duminy2018ICSC,Duminy2019FN} \end{tabular}                                                                                                   & Int.        & Yes                                                              & Yes       & Yes                                                             & Proced.                                                    & Yes \\ \hline
\begin{tabular}[c]{@{}l@{}} CHIME  \cite{Manoury2019HAI} \end{tabular}                                                                                                   & Int.        & Yes                                                              & Yes       & Yes                                                             & Planning                                                    &  \\ \hline

\end{tabular}

}

\vspace{-0.2cm}
    \caption{Comparison between the algorithms on : intrinsic vs extrinsic reward, the goal space is continuous (parametrised), single task vs multi-task problem, hierarchical learning, the action representation and whether imitation learning is used.}
    \label{table:intro:comparison}
    \vspace{-0.6cm}
\end{table}

Through this article, we have presented three implementations of a \edit{common} algorithmic framework for learning multiple control tasks through curriculum learning by discovering \edit{domain knowledge, such as the dependencies between tasks or task and action complexities,}  and exploiting this hierarchy to transfer knowledge from the easy tasks to the compositional tasks.
 Table \ref{table:intro:comparison} summarises the properties of IM-PB, SGIM-PB and CHIME in learning to perform complex tasks with compound actions, in contrast to the state of the art. While IM-PB and SGIM-PB rely on a static set of controllable and outcome features and explore the dependencies between tasks to learn sequences of actions, CHIME builds dynamically its set of tasks from emergent control models that are then used to plan sequences of actions.  Whereas IM-PB and CHIME rely only on autonomous exploration \edit{using intrinsic motivation}, SGIM-PB can request different kinds of demonstrations depending on the complexity of the target task to the appropriate teacher. All three rely on a temporally abstract representation of compound actions using task hierarchy. They efficiently manage to learn them through the discovery of relationships between tasks to enable transfer of knowledge. 
 
 We have proposed a framework unifying their common aspects: \edit{a temporally abstract representation of the relationship between tasks, learning the hierarchy of tasks with intrinsically and socially guided exploration. This summary opens a theoretical blueprint of a novel framework for robots automatic curriculum learning of uncovering domain knowledge to learn compositional tasks. We showed this can be tackled by combining hierarchical learning, planning and exploration based on intrinsic motivation and active imitation learning.}
 
In future works, we shall develop an implementation of this unified framework using all the described features: learning primitive actions and then planning sequences of them, then once learned, optimising directly these sequences owing to the procedure framework. The emergent subtasks will reduce the dependency on domain knowledge, whereas learning a representation of a compound action will result in better optimised policies and reduce the planning complexity.

\begin{acknowledgements}
This work is partially supported by the European Regional Development Fund (ERDF) via the VITAAL CPER, by Institut Mines Telecom (IMT) and by the French Ministry of Research.
\end{acknowledgements}


%
%

{\footnotesize

\bibliographystyle{spbasic}      
\bibliography{biblio}   
}

  \begin{wrapfigure}{l}{25mm} 
    \includegraphics[width=1in,height=1.25in,clip,keepaspectratio]{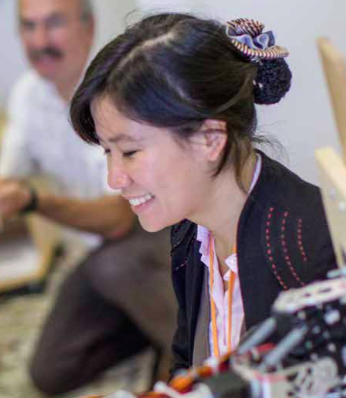}
  \end{wrapfigure}\par
  \textbf{Sao Mai Nguyen} specialises in robotic learning, especially cognitive developmental learning, reinforcement learning, imitation learning, curriculum learning for robots and human activity recognition. She received her PhD from Inria, an Engineer degree from Ecole Polytechnique and a master's degree from Osaka University, Japan. She has enabled a robot to coach physical rehabilitation in the projects RoKInter and the experiment KERAAL she coordinated, funded by the European Union FP-7 program. She is currently associate editor of the journal IEEE TCDS and co-chair of the Task force "Action and Perception" du IEEE Technical Committee on Cognitive and Developmental Systems.\par

\begin{wrapfigure}{l}{25mm} 
    \includegraphics[width=1in,height=1.25in,clip,keepaspectratio]{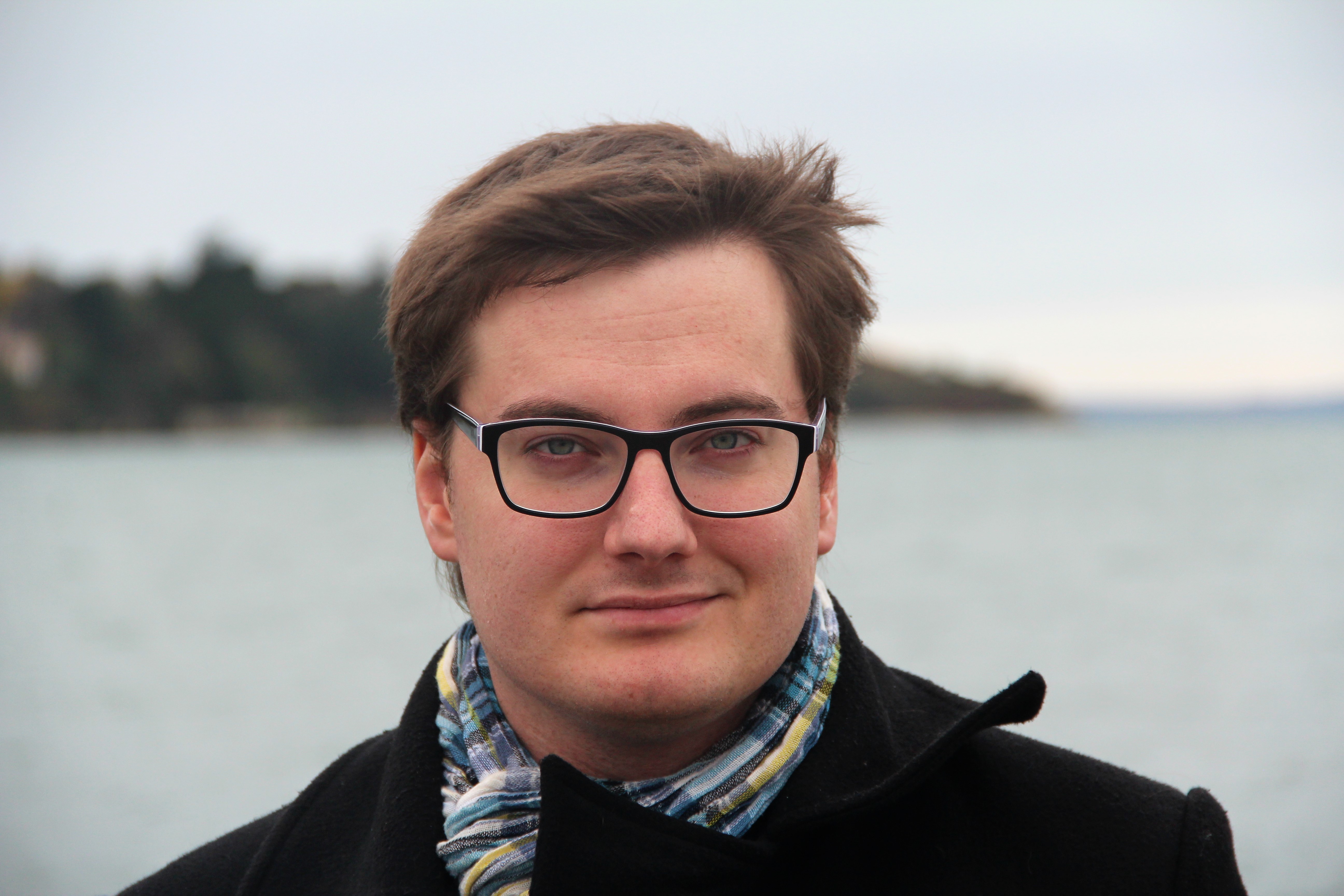}
  \end{wrapfigure}\par
  \textbf{Nicolas Duminy} holds a master in engineering from IMT Atlantique in France and has obtained in 2018 his PhD in computer science from Université Bretagne Sud in France.
His research focused on developmental robotics, and more particularly on the strategic autonomous learning of action sequences and task hierarchies.
Today, he is working as an engineer and entrepreneur to develop more immersive virtual reality experiences.\par

\par
  \textbf{Alexandre Manoury} hold a master in engineering from IMT Atlantique in France.
His research focused on developmental robotics, and more particularly on the strategic autonomous planning of action sequences and task hierarchies.
\par

\begin{wrapfigure}{l}{25mm} 
    \includegraphics[width=1in,height=1.25in,keepaspectratio]{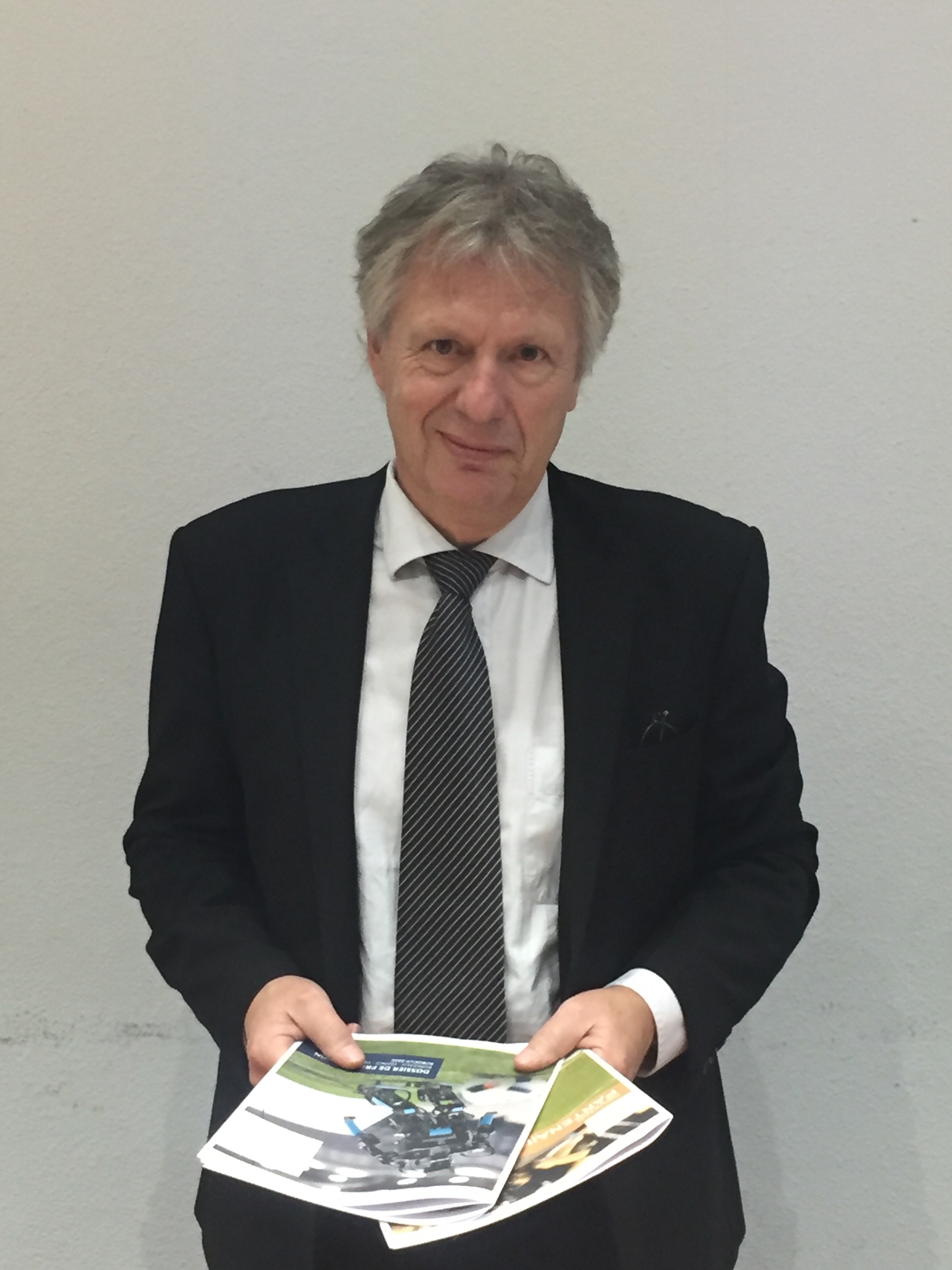}
  \end{wrapfigure}\par
  \textbf{Dominique Duhaut} received his PhD degree in computer science from the University Paris 6 in 1982 on The study of complexity of recursive function under the definition of Trahtenbrot.
He became assistant professor in 1984 in University Paris 6. In 1987, he moved to the Laboratoire de robotique de Paris where he worked on the cooperation of robots in a flexible cell.
In 1996, he was participating to the definition of the RoboCup movement that he organised in Paris in 1998. At that time his research was focused on multi-agents programming in robotics.
He moved in university of Bretagne Sud in 2000 where he became professor. His research interests are actually on self reconfigurable systems, teams of robot programming and social aspect of robots. He is also interested in science promotion for young people. He his organising competitions for schools since several years and is participating in the RoboFesta movement.\par

\begin{wrapfigure}{l}{25mm} 
    \includegraphics[width=1in,height=1.25in,clip,keepaspectratio]{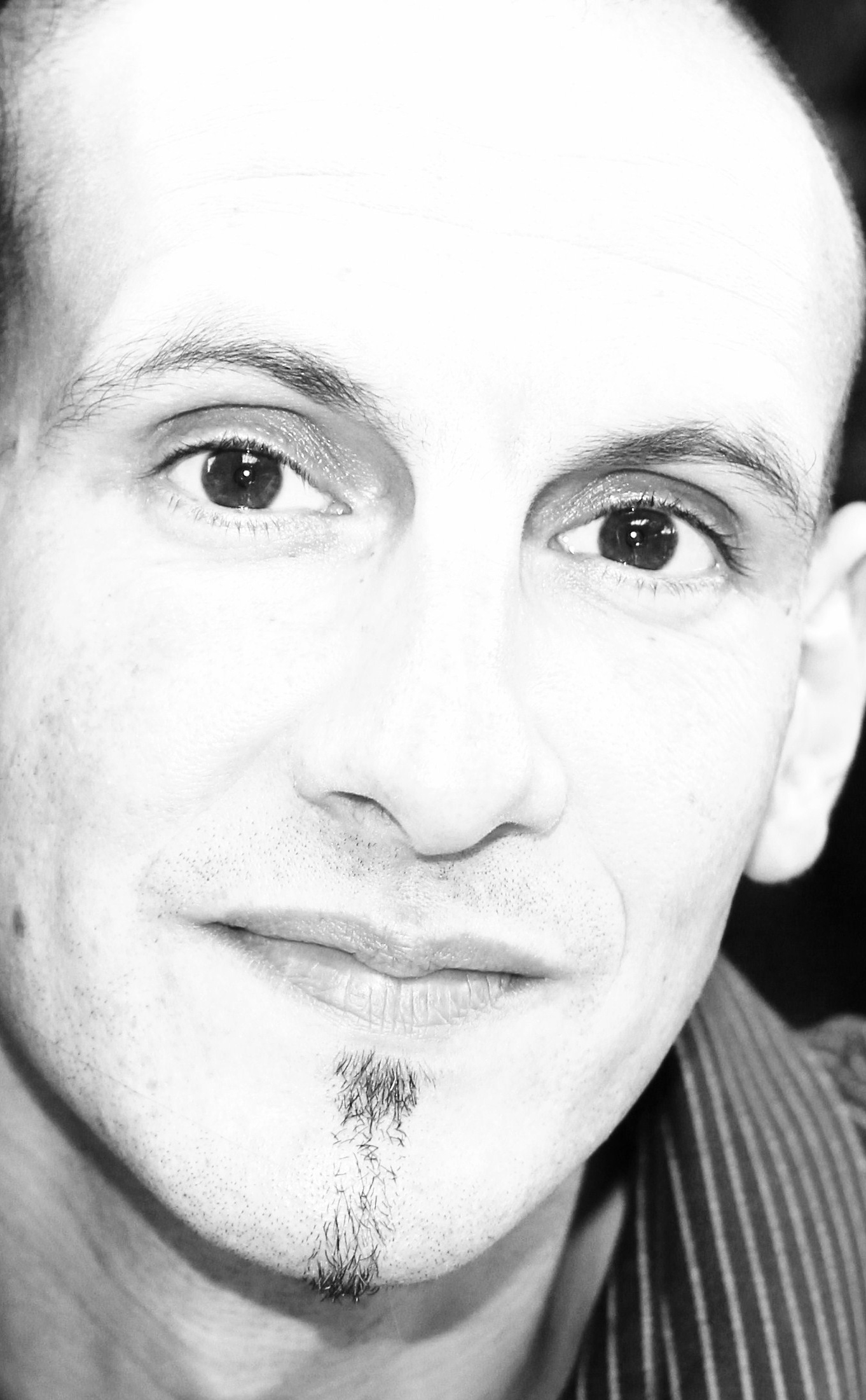}
  \end{wrapfigure}\par
  \textbf{Cedric Buche} is a professor at Ecole Nationale d'Ingénieurs de Brest in France. His research focuses on artificial intelligence and human-robot interactions. He is mainly interested in interactive machine learning approaches..\par

\end{document}